\documentclass[letterpaper, 10pt, conference]{ieeeconf}	
\IEEEoverridecommandlockouts
\newlength{\myfigwidth}
\newlength{\myfigwidthhalf}
\usepackage[dvipsnames]{xcolor}

\usepackage[bookmarks=true]{hyperref}
\usepackage{xcolor}
\usepackage{cite}
\usepackage{graphicx}
\graphicspath{ {images/} }
\usepackage{soul}

\title{\LARGE \bf
Design of a Multi-Modal End-Effector and Grasping System\\
\Large How Integrated Design helped win the Amazon Robotics Challenge
}

\author{
S.~Wade-McCue$^{\star 1,2}$,			 	
N.~Kelly-Boxall$^{\star 1,2}$,			
M.~McTaggart$^{1,2}$, 				
D.~Morrison$^{1,2}$,				
A.W.~Tow$^{1,2}$,                   
J.~Erskine$^{1,2}$, \\              	
R.~Grinover$^{1,2}$,                
A.~Gurman$^{1,2}$, 					
T.~Hunn$^{1,2}$,					
D.~Lee$^{1,2}$,                     
A.~Milan$^{1,3}$,                	
T.~Pham$^{1,3}$,                    
G.~Rallos$^{1,2}$,                  
A.~Razjigaev$^{1,2}$,\\            
T.~Rowntree$^{1,3}$,             	
R.~Smith$^{1,2}$,                   
K. Vijay$^{1,3}$,                   
Z.~Zhuang $^{1,4}$, 				
C.~Lehnert$^{2}$,                  
I.~Reid$^{1,3}$,                   %
P.~Corke$^{1,2}$,                   %
and J.~Leitner$^{1,2}$
\thanks{This research was supported by the Australian Research Council Centre of Excellence for Robotic Vision (ACRV) (project number CE140100016). The participation at the ARC was supported by Amazon Robotics LLC. Contact: {\tt\small 
sean.wademccue@hdr.qut.edu.au}}
\thanks{$^\star$S.~Wade-McCue and N.~Kelly-Boxall contributed equally to this work.}
\thanks{$^{1}$Authors are with the Australian Centre for Robotic Vision (ACRV).}%
\thanks{$^{2}$Authors are with the Queensland University of Technology (QUT).}
\thanks{$^{3}$Authors are with the University of Adelaide.}
\thanks{$^{4}$ZZ is with the Australian National University (ANU).}
}

\graphicspath{ {images/} }
\begin{document}
\maketitle
\thispagestyle{empty}
\pagestyle{empty}


\begin{abstract}
We present the grasping system behind \textit{Cartman}, the winning robot in the 2017 Amazon Robotics Challenge. The system makes strong use of redundancy in design by implementing complimentary tools, a suction gripper and a parallel gripper. This multi-modal end-effector is combined with three grasp synthesis algorithms to accommodate the range of objects provided by Amazon during the challenge. We provide a detailed system description and an evaluation of its performance before discussing the broader nature of the system with respect to the key aspects of robotic design as initially proposed by the winners of the first Amazon Picking Challenge. To address the principal nature of our grasping system and the reason for its success, we propose an additional robotic design aspect `precision vs. redundancy'.  The full design of our robotic system, including the end-effector, is open sourced and available at \href{http://juxi.net/projects/AmazonRoboticsChallenge/}{http://juxi.net/projects/AmazonRoboticsChallenge/}.
\end{abstract}

\section{Introduction}

Amazon offers approximately 400 million products to the US through their on-line marketplace\cite{ScrapeHero2017}, and is able to offer same-day shipping on many items through their Amazon Prime service.  This feat is a testament to the logistical capabilities of Amazon and showcases their state-of-the-art warehouse automation technology.  However, technological limitations have kept Amazon from entirely automating their supply chain, with the bulk of item pick-and-place tasks in warehouses still performed by humans.  Despite the strong advancement of robot and computer vision technology in recent years \cite{lin2016refinenet, levine2016learning}, pick-and-place robotics for unstructured warehouse settings is still in its infancy.  Amazon fosters development in this space by hosting an annual competition, the Amazon Robotics Challenge (ARC) (previously the Amazon Picking Challenge).

The ARC requires teams to develop autonomous warehouse manipulation systems to perform the warehouse tasks of stocking shelves and fulfilling orders into shipping boxes.  Traditional warehouses consist of static shelves in which items are stored.  In such an arrangement, travel is required when storing to and picking from the shelves.  The `goods-to-man' Kiva systems implemented by Amazon removed this requirement by having the shelves move around the warehouse autonomously\cite{Banker2009}. This design allows shelves to be packed tightly (saving floor space) and pick-and-place operations to be performed at static, distributed locations. Amazon also employs a chaotic warehouse structure where each bin in a warehouse shelf holds a large variety of items. Initially designed to improve the speed of human picking, this feature also reduces the potential for bottlenecks with peaks in item sales. Developing a system that can pick items from static, cluttered bins is the challenge presented by the ARC.

\begin{figure}[t!]
  \centering
  \vspace{-2mm}
  \includegraphics[width=\myfigwidth]{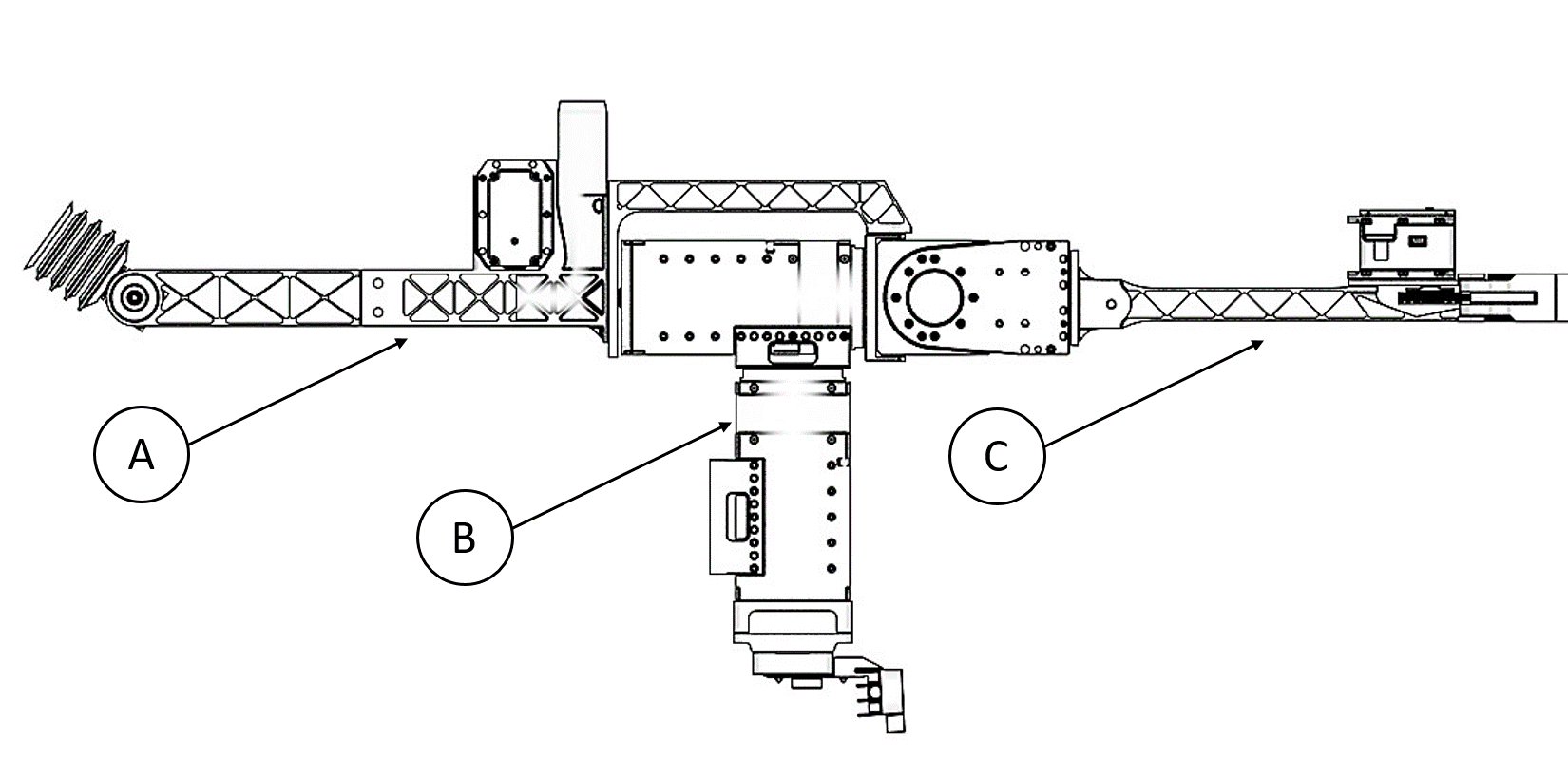}\\
 	\includegraphics[width=\myfigwidthhalf]{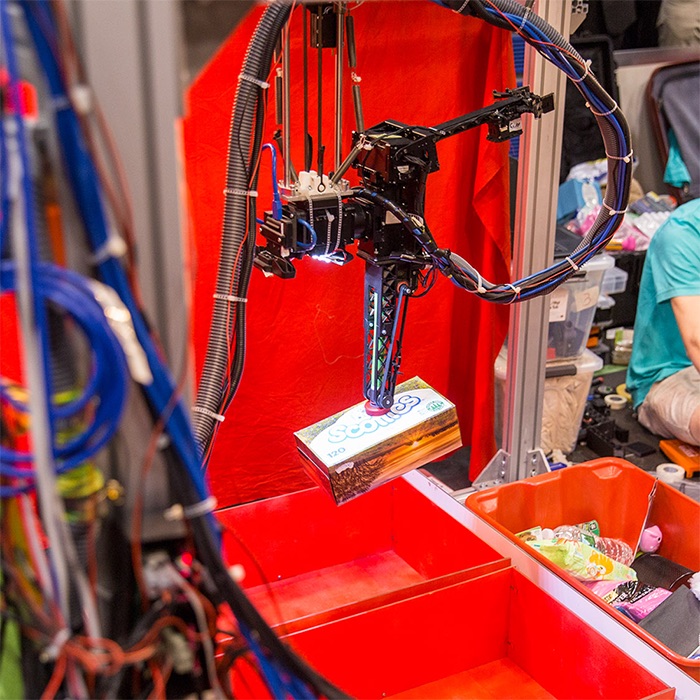}
 	\includegraphics[width=\myfigwidthhalf]{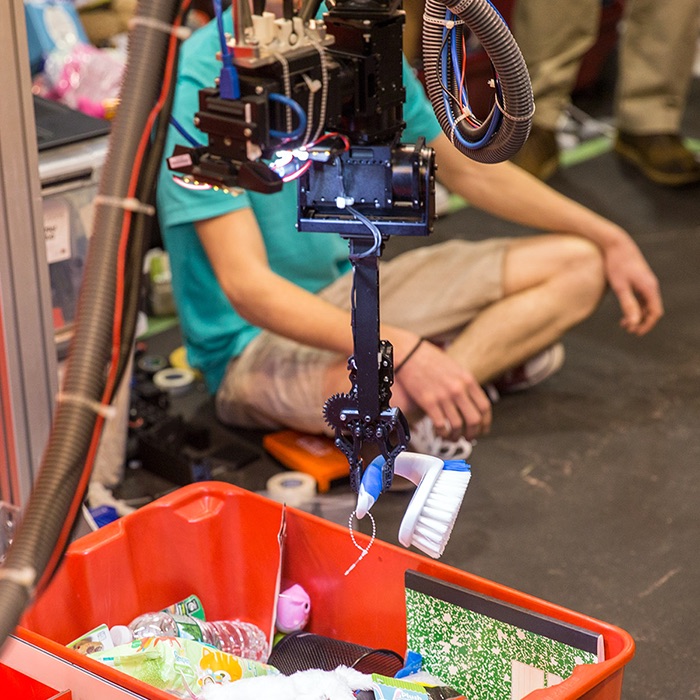}
  \vspace{-4mm}  
  \caption{Top: Model of Wrist including (A) Suction tool (B) Tool change motor and (C) Parallel Jaw Gripper, Bottom: The end-effector in use during the Amazon Robotics Challenge (left) picking an item with the suction tool and another one with the gripper (right).}
  \label{fig:toolchange}
\end{figure}


Autonomous grasping from clutter requires a robot that can handle items of varying size, weight, shape, texture and physical occlusion. On the perception front, a robot must be capable of distinguishing each item from one another. Coherent integration of both hardware and software is required to overcome the challenge of grasping items in clutter. 




We present here the grasping system of our ARC winning robot \textit{Cartman}~\cite{systems_paper}.  The grasping system consists of a hybrid end-effector with suction tool and parallel gripper, and the software, a multi-level grasp point detection algorithm designed to work with varying levels of visual information of the objects to be grasped.  The design of our grasping system was optimised for the context of the Amazon Robotics Challenge, in accordance with the four aspects of building robotic systems proposed by \cite{Eppner}: `modularity vs. integration', `computation vs. embodiment', `planning vs. feedback' and `generality vs. assumptions'.  We discuss the design of our system with respect to these aspects, and in addition propose a fifth aspect, `accuracy vs. redundancy', which allows us to more adequately represent the nature of our system within the descriptive framework.


\section{Background \& Related Work}

\subsection{The Challenge}


The 2017 ARC saw the introduction of a new set of challenges beyond previous years' competitions. A `combined' finals task was introduced and 50\% of all items in each challenge were withheld by the organisers until 45 minutes prior to the official runs. Points were awarded for successfully picking and storing objects, with bonus points awarded for successful task completion prior the assigned time limits and for handling previously unseen items. Deductions were made for damaging items or improperly recording final item locations. The stow task required robots to transfer 20 items from a tote into the storage container within 15 minutes.  The pick task required teams to pick 10 items from a hand stowed set of 32 into three cardboard shipping boxes, again within 15 minutes. Bonus points were awarded for closeable shipping boxes. The final competition combined both stow and pick with an item set of 32, and a time limit of 30 minutes. Half of the items were initially stowed, then a 10 item pick was executed from those successfully stowed items. 

\subsection{Robotic Design}
Robotic design is complex. It involves the integration of many research and engineering disciplines culminating in robust and coherent systems. After the 2015 Amazon Picking Challenge, the winning team \cite{Eppner} suggested the characterisation of robotic systems across four key spectrum-based aspects `modularity vs. integration', `generality vs. assumptions', `computation vs. embodiment', and `planning vs. feedback'.

\begin{itemize}
\item Essentially all robotic systems employ some level of integration between software and hardware. Each aspect plays a large role in overall system behaviour. 
\item Software (computation) is highly flexible and easily modified, thus  and provides opportunity for extremely rapid development of complex logical and systems
\item Hardware (embodiment) is traditionally non-flexible, yet with 3D printing technology this has changed.
\end{itemize}


\subsection{Grasping Systems}


To be successful at the ARC, a robot needs to grasp and move items between locations robustly. Grasping is a core challenge for robots~\cite{Kragic} and has seen the research and development of numerous end-effector designs, 
including suckers~\cite{Festo2015,Vaculex2017}, parallel jaw grippers (Table~\ref{GripTab}), anthropomorphic hands~\cite{Lee2016,Guo2017}, and more recently, under-actuated, compliant and soft hands~\cite{She2015}.

\begin{table}[tb]
\centering
 \caption{Open Source and Commercial Gripper Options}
 \label{GripTab}
 \begin{tabular}{p{3.5cm}p{2cm}p{1.4cm}} 
 \multicolumn{3}{c}{Different Parallel Grippers} \\
 \hline
 Gripper Name & Release Type & Drive Type \\ [0.5ex] 
 \hline
 \hline
 Adaptive Robot Gripper -  & Commercial & Servo \\ 
 Robotiq\cite{Roboti2017} & &\\
 \hline
 DHPS Parallel Gripper - & Commercial & Pneumatic  \\
  Festo \cite{Fest2017} & & \\
 \hline
 Parallel Gripper Kit A -  & Commercial & Servo  \\
 Servo City \cite{servocity} & &\\
 \hline
 Parallel Gripper \cite{Fanyeh2015} & Open Source & Stepper Motor  \\
 \hline
 DynaPincer \cite{USFRobotDesignMe2014} & Open Source & Servo  \\ 
 \hline
 Robotic Gripper \cite{Mitchnajmitch2014} & Open Source & DC Motor\\[1ex] 
 \hline
 \end{tabular}
\end{table}

Many approaches to grasp planning, including grasp point synthesis and selection, exist for most types of physical grasping systems. These methods often assume an uncluttered environment and a model that can be fitted to the object. One example of such methods is DexNet~\cite{Mahler2016}, which uses a database of 10,000 3D models to learn antipodal grasp point selection. Other grasp planning methods include but are not limited to grasp primitives \cite{Stulp2011}, shape exploration for promising grasp areas \cite{Li2015} and physics simulators such as \textit{GraspIt!} \cite{Miller2004}. However, when using suction and a parallel gripper, we have found that it can be sufficient, in fact even more competitive, to use simple heuristics and local geometric methods such as surface normals. 

In tasks such as the Amazon Robotics Challenge where many different classes of items are present, a choice between a more complex end-effector, e.g.~an anthropomorphic hand, or multiple simple end-effectors needs to be made. Multiple end-effectors require a tool change mechanism which is highly related to the manipulator used. 

There are typically two options for manipulators, one being an articulated arm \cite{UniversalRobotics2017} or a Cartesian based system \cite{Cartman_mechanism_paper}. When utilizing an articulated arm, often a hybrid end-effector is required where both attachment types are integrated into the one end-effector \cite{Yu2016} or a tool change mechanism can be used similar to a CNC Machine. A Cartesian Manipulator allows for both previous methods or a flip method where two individual end-effectors can be used in a $180^o$ offset at the wrist. For any effective and robust system it is also important to design and test the tool change mechanisms so that they can be tailored to the specific robot being used.

\subsection{Prototyping Methodologies}
Rapid prototyping is considered central to the success of countless disruptive innovations within both research and industry. Kelley and Littman~\cite{Kelley2004} describe the innovative prototyping behind the success of Amazon and Apple, essentially presenting it as the process of problem solving without wasting time. We elaborate on this definition, proposing that rapid prototyping is the process of \textit{meaningful} problem solving, executed over a \textit{minimum} time. Methods applicable to robotics development which facilitate this definition include maximizing system uptime during testing, 3D printing for hardware development and improved development team productivity\cite{Hudson2007,Bennis1997}.


\section{System Description} \label{Description}


Here we provide an overview of our grasping system as implemented on \textit{Cartman} \cite{systems_paper}, the winning Amazon Robotics Challenge robot. The grasping system takes semantically segmented RGB-D images from the vision system~\cite{Cartman_vision_paper}, synthesizes and selects grasp points, then physically attaches and detaches from items using either the suction tool or the parallel gripper. In the following, we describe the end-effector hardware, software and their integration. The role of redundancy in the complimentary solutions is explored throughout, while linking with the functionality of the system. It is shown that the strategic integration of sub-systems provides increased robustness, which was key to a winning grasping system.

\subsection{Hardware}
Our grasping solution implements both a suction tool and a parallel gripper, both of which are mostly 3D printed using PLA, a common and easily extrudable bioplastic. The Cartesian mechanism \cite{Cartman_mechanism_paper} combined with the wrist motors (Fig. \ref{fig:toolchange}, B)  allow for 6DOF Cartesian pose control of each tool, thus simplifying trajectory planning. Suction is \textit{Cartman}'s primary tool, as initial trials found that most items in the Amazon item set were more easily sucked than gripped.

\subsubsection{Suction Tool}
\begin{figure}[tb]
  \centering
  \includegraphics[width=\columnwidth]{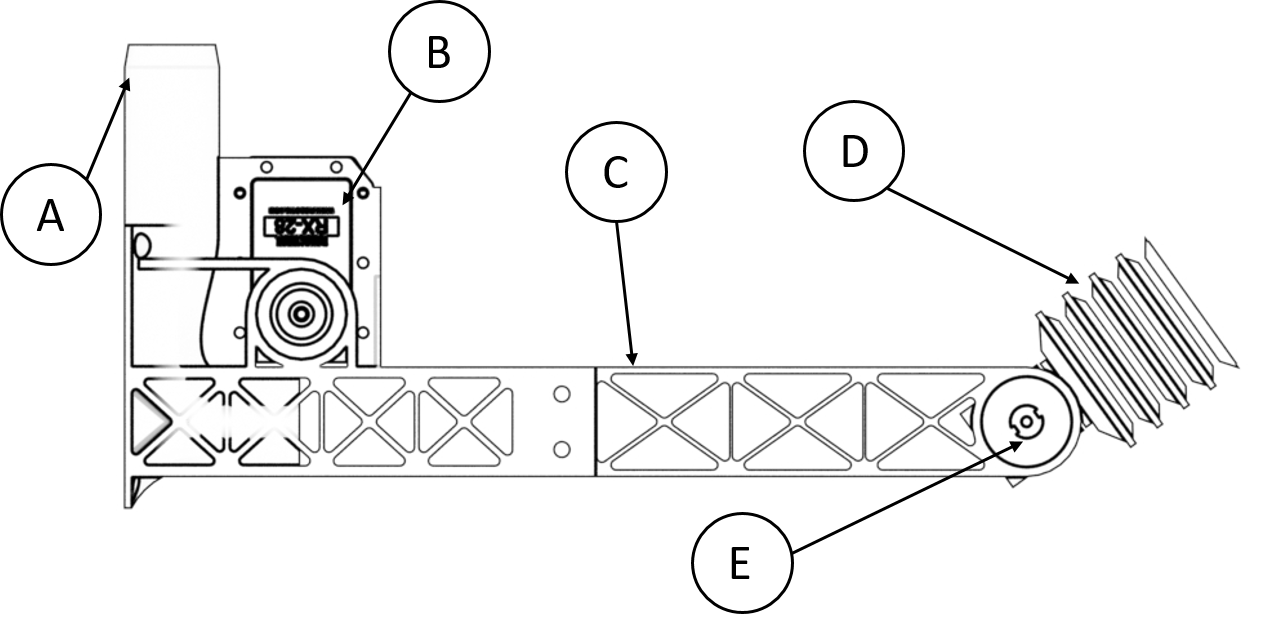}
  \caption{Model of Suction Tool, (A) Vacuum Inlet, (B) Powered by a Dynamixel RX-10, (C) 3D printed body, (D) 40mm silicone suction cup and (E) Belt-driven joint for 6DoF articulation. }
  \label{fig:suction}
\end{figure}

\begin{figure}[b]
  \centering
  \includegraphics[width=\columnwidth]{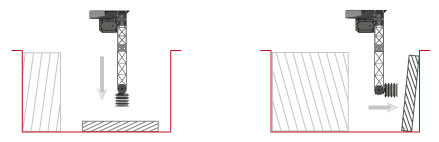}
  \caption{Example of why articulation at the end of the suction tool was useful.}
  \label{fig:suctionex}
\end{figure}

A key feature of our suction mechanism (Fig. \ref{fig:suction}) is the articulation (label E) available close to the tool tip (label D), similar to Team NimbRo's solution~\cite{schwarz2017apc} for the 2016 APC. This belt driven mechanism allows for low profile vertical entry into cluttered scenes while grasping vertical surfaces (Fig. \ref{fig:suctionex}). A 40mm silicone suction cup was found to grasp most smooth and non-porous items successfully, providing enough lifting force for the heaviest items, weighing up to 1kg. To achieve this result the suction is driven by two vacuum pumps in parallel, each rated with a free air flow rate of 10CFM and an ultimate vacuum pressure of 0.3Pa.


\subsubsection{Parallel Gripper}
\begin{figure}[t]
  \centering
  \includegraphics[width=3in]{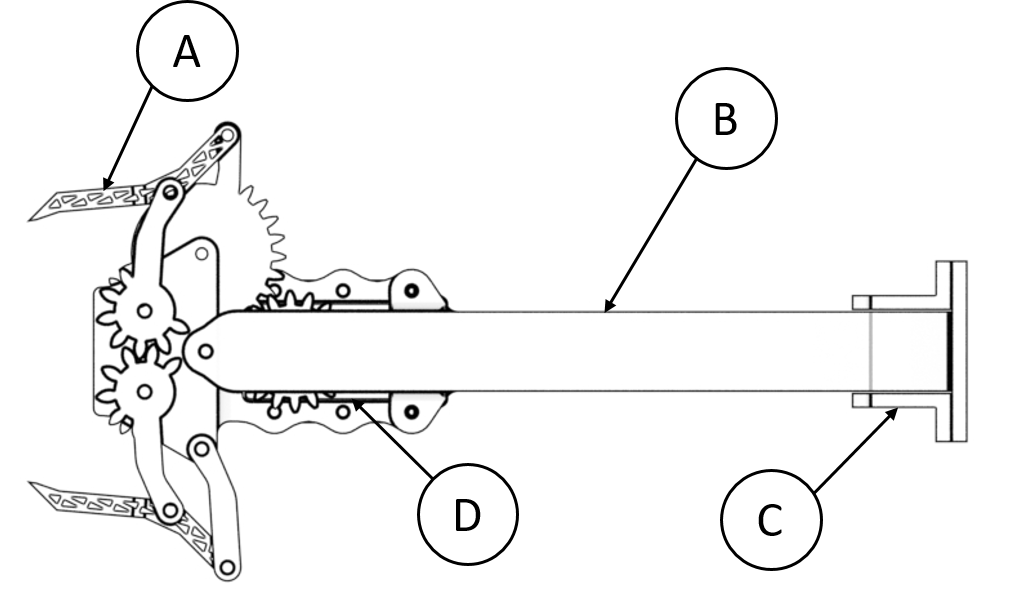}
  \caption{Model of Parallel Gripper, (A) Gripping plate with design for maximum strength and minimum weight, (B) Gripper extension arm, (C) Quick swap bracket and (D) Drive system from Dyanmixel RX-10.}
  \label{fig:gripper}
\end{figure}

Our gripper (Fig. \ref{fig:gripper}) is specifically designed to compliment the grasping capabilities of the suction tool within the Amazon Robotics Challenge item set. Doing so limits the number of items that the gripper is required to grasp, thus facilitating a more targeted design. Given that the sucker is capable of grasping most of the larger items, this led to a more compact and unobtrusive gripper. It utilizes a simple four-bar linkage and gear mechanism driven by a Dynamixel RX-10 to actuate the gripping plates. Each plate is angled inwards by $10^o$ to provide a tighter pinch at the tip and rubberized tape is applied to the plates to provide compliance.

\begin{figure}[b]
  \centering
  \includegraphics[width=3in]{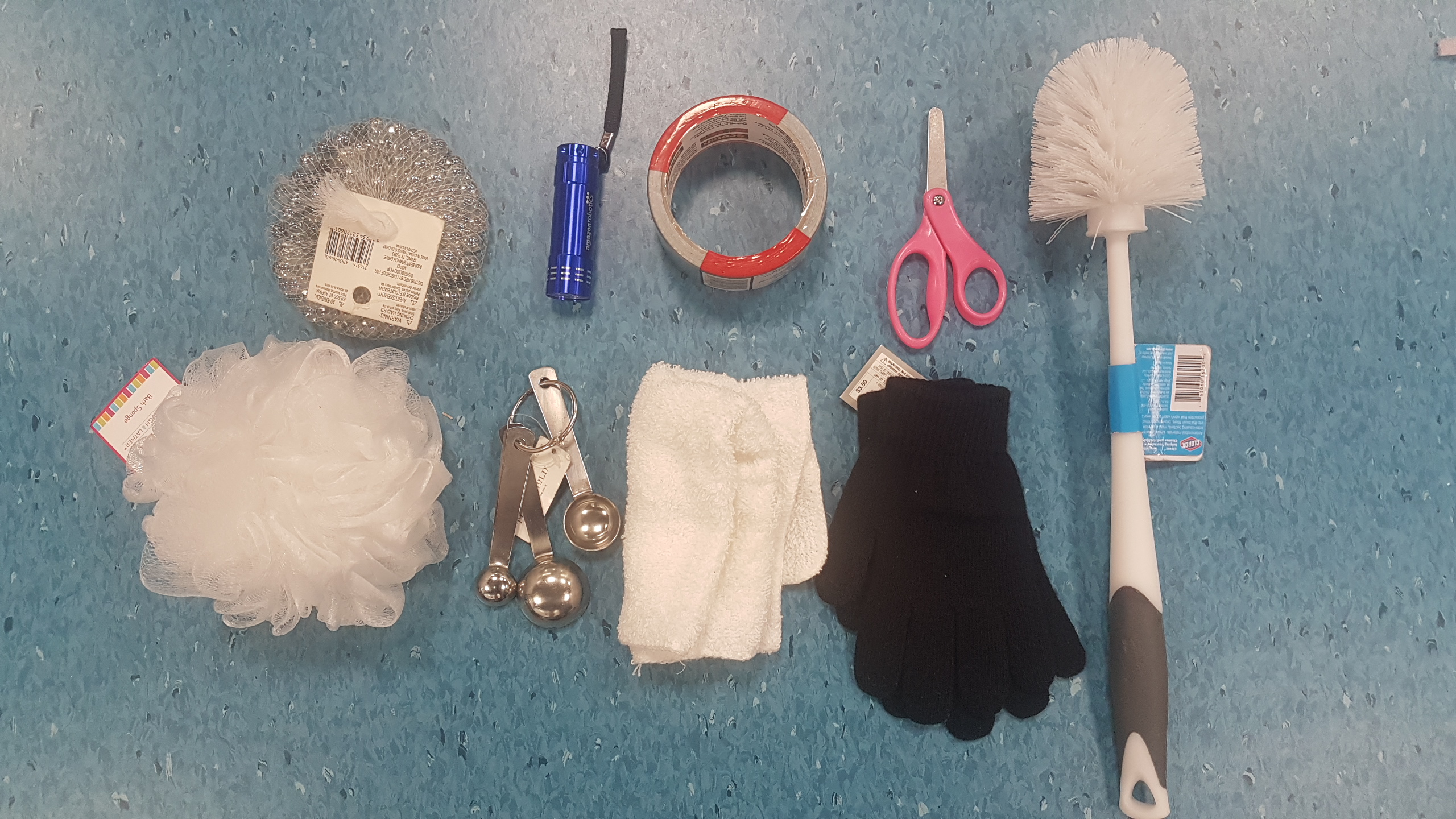}
  \caption{Items that could not be picked via suction with our system.}
  \label{fig:gripitems}
\end{figure}

\subsubsection{Tool Selection Mechanism}
The gripper and sucker are mounted $180^o$ either side of the wrist assembly (Fig. \ref{fig:toolchange}). The arrangement of servo motors on the wrist allows for a simple rotation of the Tool Change Motor (B) to determine which tool is in the active position, pointing down toward the storage system. The procedure is simple and rapid, thus avoiding complex mechanisms and saving valuable competition time. This mechanism also aided in the prototyping phase where either the sucker or gripper design could be changed significantly without affecting the design of the other.




\subsection{Software} \label{graspsynthesis}
The software behind our grasping system incorporates grasp point synthesis, grasp point selection and tool selection.

\subsubsection{Grasp Synthesis}
\begin{figure}[t!]
  \centering
  \includegraphics[width=3in]{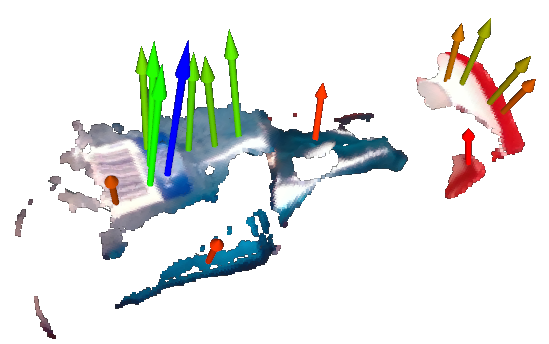}
  \caption{Grasp point hypotheses generated on an object using surface normals. The longer green arrows indicate more likely grasp points and the blue arrow denotes the best proposed candidate.}
  \label{fig:grasppoint}
\end{figure}

We use a pixel-wise semantically segmented RGB-D image \cite{Cartman_vision_paper} as the input to our grasp synthesis algorithms. This provides the system with information about the identity and pixel-accurate location of items in its view (cf. Fig.~\ref{fig:grasppoint}). 

We use three approaches to grasp synthesis, \textit{surface normals}, \textit{RGB-D centroid} and \textit{RGB centroid}. The \textit{surface normals} approach uses the raw point cloud segment of the target item. After generating surface normals for each point, heuristics such as distance from the segment edge and angle from vertical are used to prune undesirable normals. Finally, surface normals are ordered such that two desirable poses are positioned spatially far from one another to allow for multiple grasp attempts on the same item in meaningfully different locations. Example grasp points from the surface normal approach can be seen in Fig. \ref{fig:grasppoint}. The \textit{RGB-D centroid} approach chooses the centroid of the segment as the grasp pose. Here the nearest available depth point is chosen if one at the centroid is unavailable. The orientation of the pose is selected to be vertical. The \textit{RGB centroid} approach differs from the RGB-D by assuming the centroid's depth. This assumed depth is used to compute the world x and y position so that the tool point can be positioned above the item. As a grasp is executed, the arm is lowered slowly into the scene, receiving feedback when to stop from the scales beneath each bin, or a flow sensor in the case of suction.
Across all three grasp selection approaches, principal component analysis (PCA) is used to orientate the tool point with the principal axes of the target item. This allows the gripper to choose antipodal points and provides an estimation of the item's pose to aid in packing. We found this method to work robustly for grasping a variety of items; the approach became erroneous when an item was occluded or a poor segment was received from the semantic segmentation system.

The selection of \textit{surface normals}, \textit{RGB-D centroid} or \textit{RGB centroid} is decided based on the quality of the RGB-D image returned for that item. The default method is \textit{surface normals}, unless the item requires gripping or the RGB-D image is too noisy to provide clean surfaces. For items which give valid depth points but have indistinguishable surfaces (such as the bath sponge), we use \emph{RGB-D centroid}. Finally, \emph{RGB-Centroid} is used when no depth information is recoverable from an object. This is the case for black and highly reflective or transparent objects. Here, we use weight sensors installed under the storage system to detect collision with the object and to begin attachment.\\
All grasp classes employ principal component analysis (PCA) to align the end-effector with the major axis of the item. This is important for two reasons, firstly the gripper requires alignment with long items (such as the torch and toilet brush) to successfully grasp around the circumference of the shaft. Secondly, this allows the robot to align items when packing them into the storage system or shipping boxes.


\subsubsection{Grasping Class} \label{class}
By combining the grasp synthesis with tool selection, it can be said that each item belongs to a grasping class, which defines the entire approach that the grasping system takes for any given item, including which grasp synthesis algorithm and which tool is used when manipulating it.
Each item belongs primarily to one of the five grasping class listed below:
\begin{itemize}
\item Surface-normals, suction
\item RGB-D-Centroid, suction
\item RGB-D-Centroid, grip
\item RGB-centroid, suction
\item RGB-centroid, grip
\end{itemize}

\section{Evaluation of performance}

Our end-effector must be capable of grasping items in both cluttered and uncluttered environments. We evaluate the performance of both the gripper and the sucker in uncluttered environments separately, before evaluating our end-effector as a whole on both cluttered and uncluttered environments.

\begin{figure}[t!]
  \centering
  \includegraphics[width=3in]{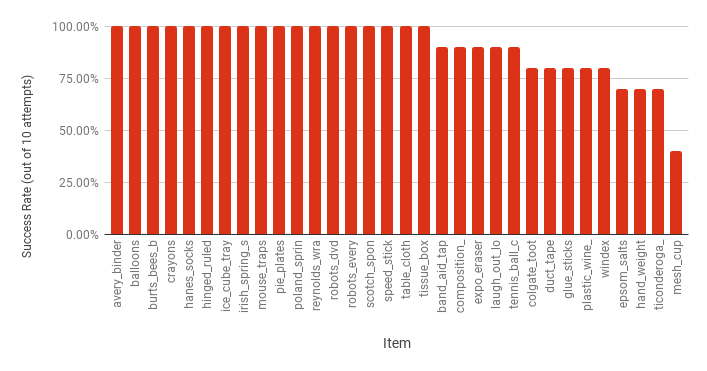}
  \caption{Picking success of the suction tool in an uncluttered environment}
  \label{fig:suckdata}
\end{figure}

The suction tool is evaluated in cluttered scenes by attempting to grasp each item ten times across a variety of orientations, see Fig. \ref{fig:suckdata}. Items that were incapable of being sucked were instead evaluated using the gripper. All grasp attempts employed the vision pipeline used in the competition system; please refer to~\cite{systems_paper, Cartman_vision_paper} for a detailed description of the competition system and the semantic segmentation approach, respectively.

The suction tool is used for thirty-two of the seen items and was found to be robust at picking most. Fig.~ \ref{fig:suckdata} shows the picking-by-suction success rate across all items in question. The system struggled on the \textit{mesh cup} where suction was only possible in two item orientations where the base was facing upwards or downwards. The sucker also struggled on the \textit{hand weight} and \textit{epsom salts}, the heaviest items in the set. The trade-off between suction cup size and robustness to all objects lead to high success rates even on the heavier items. Suction grasps most commonly failed when the target item was positioned in an unstable orientation, i.e. resting against a wall (cf. Fig. \ref{fig:suctionex}, right).

\begin{figure}[b]
  \centering
  \includegraphics[width=\columnwidth]{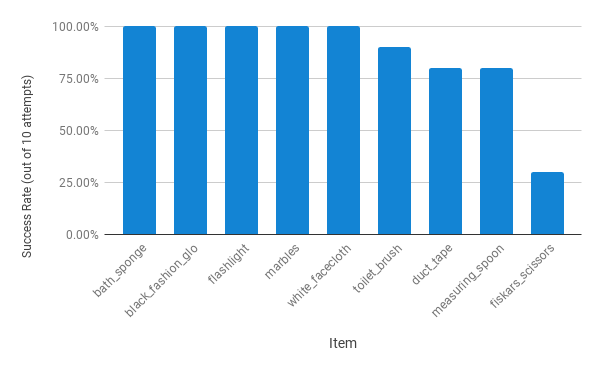}
  \caption{Picking success of the gripper tool in an uncluttered environment }
  \label{fig:gripdata}
\end{figure}

The gripping tool is used for nine of the training items categories. Fig.~\ref{fig:gripdata} shows success rates for picking items using the gripping tool. As can be seen it is robust to picking all items except for the \textit{fiskars scissors}. The poor performance on \textit{fiskars scissors} resulted from poor segments by the semantic segmentation approach, which struggles with small and thin objects with shiny surfaces. The segments produced often captured the handles only, rather than the blades which are more ideal for antipodal grasps. While the performance here was low, the system was able to pick the scissors every fourth attempt on average.

\begin{figure}[t]
  \centering
  \includegraphics[width=\columnwidth]{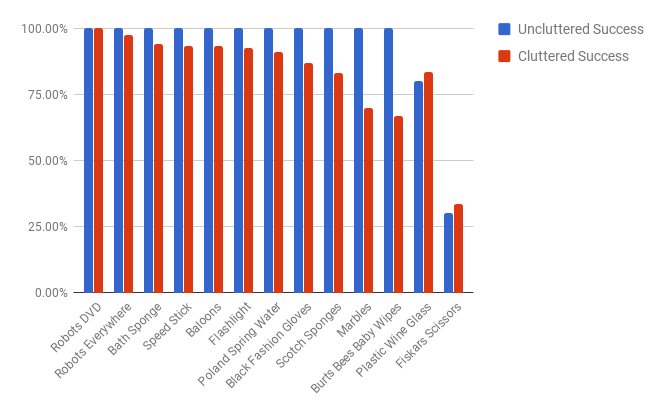}
  \caption{Success rates of items in uncluttered vs cluttered environment }
  \label{fig:comparison}
\end{figure}

The end-effector is tested on cluttered scenes over a 7 hour period in a finals-style task. The objective is to move a set of items between the stow tote and storage bins repeatedly over a long time period. Seventeen items that represented a wide cross-section of the competition items are used. The seventeen items ticked each of the following physical aspects: rigid, semi-rigid, deformable and hinged. The items also ticked the following visual aspects: opaque, transparent, partially transparent, reflective and IR-absorbing. Cluttered scene grasping performance is compared against corresponding uncluttered success in Figure \ref{fig:comparison}. Note the reported statistics do not include grasp failures that resulted from misclassification.

The comparison between cluttered and uncluttered grasping performance highlights that the end-effector design and grasp synthesis system coped well in both environments with the system having a $90\%$ success rate in the uncluttered environment and $83\%$ on the seen items in a cluttered environment. We argue this performance is a direct result of the multiple levels of redundancy designed into the system by having multiple grasp synthesis approaches and two grasping tools to select between. 

Large differences in performance between cluttered and uncluttered environments were easily explained. For example, the \textit{Burt's Bees Baby Wipes} experienced issues in cluttered scenes when occluding items added additional weight to the already heavy item. A large number of grasps also fell on its lid that was prone to opening, introducing a shock load that rippled the item out of grip.

\section{Design Aspects} \label{design}
We now discuss our system under the framework originally proposed by \cite{Eppner}, and extend the framework to include a new aspect, precision vs. redundancy.

\subsection{Modularity vs. Integration}
Modular systems allow for a decomposition of complex systems into discreet functional units which, when interfaced together, build a complete system. Integration allows for the seamless operational performance of subsystems to build a complete system. \textit{Cartman} and its grasping system are highly modular and integrated systems. Where modularity aided in the design process, the integration of complimentary systems such as the sucker and gripper greatly improved performance and increased redundancy. Further, the parallel development of hardware and software components led to highly integrated solutions stemming from both the hardware and software. This added to overall robustness without committing to unnecessarily complex solutions to problems that could be solved with greater integration and redundancy.

\subsection{Computation vs. Embodiment}
Computation and embodiment are equally essential to the success of our grasping system. We aimed to use each one to reduce the development load on the other. When faced with a performance barrier or design problem, computation and embodiment were employed in tandem to afford the most effective solution accounting for performance gained vs. development resources required. This was made possible largely due to our mechanical design capabilities through the use of 3D printing, thus allowing the embodiment to update with or often instead of computation. 

\subsection{Planning vs. Feedback}

The grasping system used a minimum of both planning and feedback to maintain a low overall complexity. 
Sophisticated feedback such as high resolution torque sensing and tactile feedback can drastically increase the price of hardware, thus reducing accessibility to the technology. The Cartesian manipulator reduced planning complexity when compared to that of a typical robotic arm. Furthermore, our movements were point-to-point that planning was simple and often only required straight line movements. 

\subsection{Generality vs. Assumptions}

Our grasping system utilises a combination of specific and available solutions, which when implemented in parallel show a more generalised grasping capability. For instance, any single grasping class from Section \ref{class} is a specific solution which works for only a subset of items. However, the combination of all the classes when applied to items appropriately allows our system to successfully grasp all of the Amazon items. Regarding hardware, we used two less-general solutions (sucker and gripper), which when combined improved the generality of the overall system.

\subsection{Precision vs. Redundancy}
Robotic systems necessarily operate with a trade off between precision and redundancy. For instance, precise systems may operate with few or no levels of redundancy to provide a similar overall reliability compared to a highly redundant, yet less precise system. We propose that there are important considerations to be made within this design spectrum. Precise systems often require highly accurate sensors and control systems, and are thus more expensive to produce. In some instances this in unavoidable, such as in manufacturing where precise end-effector movement is crucial. By contrast, highly redundant systems make use of cheaper sensors and control hardware where possible by instead leveraging performance gained from multiple layers of redundant systems. The overall design of \textit{Cartman} \cite{systems_paper} and its grasping system is a prime example of a robotic system that makes use of redundancy to minimise the impact that less complete subsystems or precise components may otherwise have on performance. This is observed in both the hardware and the software which provide flexibility by allowing a range of grasping methods with complimentary performance.

\section{Conclusion and Future Work}
Effectively combining multiple solutions to the same problem increases system robustness considerably. In light of this, we propose the introduction of a fifth robotic design aspect; `precision vs. redundancy'. Against this design aspect, our solution is weighted strongly towards redundancy; reducing reliance on accurate and costly hardware while maintaining overall robustness. 
We have demonstrated the effective use of robotic design principals, culminating in a robust, multi-modal grasping system capable of successfully completing $90\%$ and $83\%$ of grasps in uncluttered and cluttered environments respectively. When implemented within \textit{Cartman}, this system won the 2017 Amazon Robotics Challenge.

Use of a multi-level grasp synthesis approach means time is potentially wasted trialling unsuitable grasp selection approaches. This problem was tackled in the competition through manual assignment of grasping classes to items. Learned or autonomous item grasp type classification is a promising avenue for improving the applicability of \textit{Cartman} to a larger variety of items. 


\addtolength{\textheight}{-17cm}   

\bibliographystyle{IEEEtran}
\bibliography{grasping}

\end{document}